\documentclass[a4paper, 10pt, conference]{ieeeconf} 
\AtBeginDocument{\let\autocite\cite}

\IEEEoverridecommandlockouts                              

\usepackage{balance}
\usepackage{url}
\usepackage{cite}
\usepackage{graphicx} 
\usepackage{flushend}
\usepackage{booktabs}
\usepackage{tikz}
\usepackage{amsmath}
\newcommand\copyrighttext{%
  \footnotesize \textcopyright 2026 IEEE.
  Permission from IEEE must be obtained for all uses, in any current or future
  media, including reprinting/republishing this material for advertising or promotional
  purposes, creating new collective works, for resale or redistribution to servers or
  lists, or reuse of any copyrighted component of this work in other works.}
\newcommand\copyrightnotice{%
\begin{tikzpicture}[remember picture,overlay]
\node[anchor=south,yshift=10pt] at (current page.south) {\fbox{\parbox{\dimexpr\textwidth-\fboxsep-\fboxrule\relax}{\copyrighttext}}};
\end{tikzpicture}%
}

\begin{document}
\IEEEoverridecommandlockouts
\overrideIEEEmargins

\title{\LARGE \bf
  Feasibility of Embedded Photoplethysmography Sensing in Short-Duration Tactile Interactions With Pocket-Sized Robots Using IMU- and Confidence-Based Filtering
}

\author{Turjja Datta$^{1}$, and Morten Roed Frederiksen$^{2}$
  \thanks{{$^{1}$Turjja Datta and $^{2}$Morten Roed Frederiksen\tt\small mrof@itu.dk} are affiliated with the Data Systems \& Robotics Department of The IT-University of Copenhagen.}
}

\maketitle
\copyrightnotice
\begin{abstract}
Ubiquitous companion robots offer a promising avenue for immediate anxiety relief in children, yet their effectiveness relies on the ability to monitor physiological states continuously and unobtrusively. Current solutions often depend on external wearables, which impose usability barriers and limit the robot's autonomy. This paper investigates the integration of an embedded photoplethysmography (PPG) sensor directly into a pocket-sized companion robot, AffectaPocket, to enable self-contained heart rate monitoring during tactile interaction. We address the significant challenge of motion artifacts inherent in handheld usage by implementing a two-stage filtering pipeline that utilizes an onboard Inertial Measurement Unit (IMU) to reject high-variance segments and a confidence-based smoothing algorithm for recovery periods. We evaluated the system against a commonly used wrist worn sensor in a Within-Subjects Study with 26 participants. Our results demonstrate that the filtering strategy significantly reduced the Mean Absolute Percentage Error and achieved statistical equivalence to the ground truth measurements (p$<$0.05). Analysis of short-duration interactions shows that the sensor requires stability over longer periods to converge.
\end{abstract}
\section{Introduction}

Anxiety disorders represent a significant public health challenge in the Western world, with 11\% of children in the US receiving a formal anxiety disorder diagnosis before the age of 17 \cite{CDC_Childrens_Mental_Health_2025, Ghandour2019}.
To investigate the impact of using robots as a support device, we developed \textit{AffectaPocket}, a pocket-sized ubiquitous robot designed with the intention to mitigate acute stress by providing tactile grounding and diverting attention away from anxiety-inducing stimuli \cite{Frederiksen2024TactileCL, Frederiksen2024TowardAP}.

Previous evaluations of AffectaPocket's efficacy relied on external heart-rate monitors to track physiological responses \cite{Schaefer2014AFS, Frederiksen2024TactileCL}. While these wearable devices offer high-precision continuous measurement, they impose significant usability burdens on children, including correct placement and connectivity management \cite{McElwain2023EvaluatingUE, app15073512}. Integrating biometric sensing directly into the robot represents an important step toward future closed-loop biofeedback. However, such closed-loop use requires reliable temporal resolution and robustness during natural interaction. This paper investigated the feasibility of using embedded sensing to provide usable heart-rate estimates during handheld tactile interaction \cite{app15073512, Talbot2021AdvancesIA, 10.1016/j.intcom.2008.10.011}.

Embedding optical sensing into a handheld, pocket-dwelling device presents technical challenges. The confined environment of a child's pocket compromises stable sensor-skin coupling, and the robot's primary interaction modality (tactile squeezing) introduces severe motion artifacts that degrade the reliability of standard photoplethysmogram (PPG) signals \cite{electronics3020282, Faria2025AdvancingEA}.
\begin{figure}[h]
\centering
\includegraphics[width=0.50\textwidth]{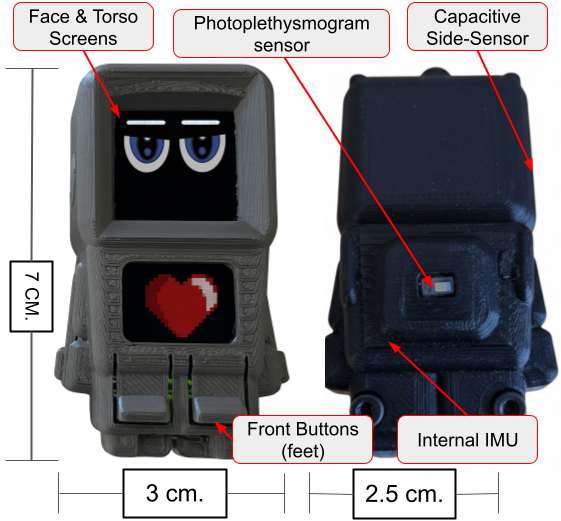}
\caption{The AffectaPocket ubiquitous robot used in the experimental study. Designed to offer thought diversion for children with anxiety disorders through tactile grounding, the device is intended to be held and squeezed. The posterior view (shown) reveals the embedded photoplethysmogram (PPG) sensor used to estimate the user's heart rate during these dynamic interactions.}
\label{fig:affecta}
\end{figure}

Reliable biosensing and dynamic state estimation have long constituted a fundamental challenge in Human-Robot Interaction (HRI), particularly as systems transition from controlled laboratories to the unstructured reality of the real world \cite{Kocielnik2013SmartTF, Qiao2022ImprovingPO}. While substantial research has focused on robot perception of human affect through external cameras \cite{4468714, Ahmad2022} or voice analysis \cite{Wieringa2005ContactlessMW}, the field of in-hand physiological sensing remains underexplored. Unlike stationary medical devices or tightly strapped wearables \cite{lee2020, Yin2024WearableAI}, interactive robots are subject to erratic manipulation, varying pressure, and unpredictable orientation \cite{7759417, doi:10.1177/02783649211050958}. Enabling accurate physiological measurement under these dynamic conditions is critical for the next generation of socially assistive robots, but the present work treats heart rate as a physiological sensing target that may support future affective interaction, rather than as a validated standalone measure of emotional state \cite{9502508}.

This paper proposes a hardware and software solution for improving heart-rate estimation in pocket-sized tactile robots under mechanical constraints. We expanded the \textit{AffectaPocket} architecture to include an embedded PPG sensor and an Inertial Measurement Unit (IMU). To address signal noise, we developed a multi-stage filtering algorithm that uses IMU data to detect motion-heavy segments and applies a confidence-based smoothing strategy during post-motion recovery periods. The system was validated in a within-subjects study ($N=26$), where the robot's onboard measurements were compared against reference data simultaneously captured by a Polar Verity Sense optical heart-rate sensor. The study evaluates the reliability of embedded heart-rate estimation during tactile interaction; it does not evaluate continuous emotional-state recognition or anxiety classification.

Our study primarily hypothesized that mechanical artifacts inherent to a pocket-sized, handheld robot would degrade optical heart-rate estimation, but that a multi-stage filtering strategy based on accelerometer variance and confidence smoothing could recover a signal suitable for longer-duration trend monitoring. The results support this feasibility-oriented framing: the proposed filtering pipeline significantly reduced the Mean Absolute Percentage Error (MAPE) from 29.14\% to 22.72\% and achieved statistical equivalence with the reference sensor within $\pm 5$ BPM bounds. However, the system remained limited for single-measurement or short-duration state estimation, achieving only 48\% accuracy when classifying heart-rate zones. We therefore position the contribution as an incremental but meaningful systems extension of \textit{AffectaPocket}: embedding PPG sensing into the robot itself and demonstrating that IMU- and confidence-based filtering can improve aggregate sensing quality during tactile interaction, while identifying remaining barriers to reliable instantaneous feedback.

\section{Related Work}

\subsubsection{Advances in Therapeutic Robots}
In response to systemic bottlenecks where waiting times for professional assessment can be prolonged, socially assistive robots (SARs) are increasingly explored to support emotional well-being \cite{Hofstede2025AFS, Trost2019SociallyAR} and reduce pain or anxiety \cite{Trost2020SociallyAssistiveRU, Lima2025PromotingCH, Carnevale2025ExploringTI}. These robots often use tactile and social interactions \cite{Frederiksen2024TowardAP, Shiomi2025DifferentialEO}, cuddly textures \cite{Bates2020SnuggleBotAN, Shiomi2025WhatMA, Shahab2024ManufactureAD}, breathing motions \cite{Matheus2022ASR, Terzioglu2020DesigningSC}, coupled with additional cues to comfort users \cite{Matheus2024OmmieTD}. For example, Paro the seal \cite{Batt2025AnEO, Granier2023GettingTS} responds to petting and emits soothing sounds, and Kaspar \cite{Milling2022InvestigatingAS} (a child-sized humanoid) is equipped with skin sensors so it reacts when touched, encouraging children with autism disorders to share emotions and take turns \cite{Robins2018KasparTS}. Plush companions have similar designs: Purrble is a small stuffed “anxious creature” that purrs with a rapid heartbeat when distressed and slows as the child soothes it \cite{Williams2024}, and there are a few research projects that focus on hugging as a unique interaction between humans and robots \cite{Goris2011MechanicalDO, Saldien2006ANTYTD, Goris2008TheHR, Block2022InTA}. 

\subsection{Bio-Sensing and Signal Filtering Robots in HRI}
While therapeutic robots have demonstrated behavioral efficacy, the field is undergoing a paradigm shift from socially interactive to physiologically adaptive systems that employ embedded bio-sensing to objectively measure and regulate user states in real-time \cite{Liu2006AffectiveSR, Rani2007AnxietybasedAC, Bolano2021DesignAE, 10.1016/j.intcom.2008.10.011}. Addressing the limitations of external camera-based analysis—such as occlusion during hugging and the subjectivity of facial expressions—modern platforms integrate Photoplethysmography (PPG) \cite{electronics3020282, Rihet2024RobotNI} and Electrodermal Activity (GSR) sensors \cite{Hald2020HumanRobotTA, Lu2025EffectsOA} directly into the robot's shell to capture autonomic nervous system metrics like Heart Rate Variability (HRV) \cite{Malik1996HeartRV, Shao2021ComparisonAO} and skin conductance \cite{Pei2024TowardOA,Savur2023SurveyOP, Rosli2022EnhancingSR, Critchley2002ElectrodermalRW}. There have also been efforts into standardizing the processing of these signals \cite{Koelstra2012DEAPAD, GmezLara2019FeatureEA, Roy2020}. Examples of such is the HRI Physio Lib, a modular framework that facilitates the closed-loop biocybernetic adaptation of robot behaviors based on real-time physiological classification \cite{Kothig2020HRIPL, Chi2023ASO}. However, the deployment of these systems in naturalistic settings can be challenged by motion artifacts (MA) caused by mechanical deformation during interaction, which can obscure otherwise beneficial optical PPG signals \cite{Allen2007PhotoplethysmographyAI, Rihet2024RobotNI}. Recent research has also demonstrated that adaptive filtering using Least Mean Squares (LMS) algorithms and accelerometer-based reference signals can help to reduce heart rate estimation errors \cite{Poh2010MotiontolerantME, Zhang2014TROIKAAG, Zhu2015MICROSTAM}. Furthermore, to address the short duration of therapeutic interactions, which often precludes standard spectral analysis, researchers have validated the use of ultra-short-term windowing (10–30 seconds) and event-locked filtering strategies to reliably detect transient micro-states of stress and engagement \cite{Muoz2015ValidityO, Gallardo2021HeartRV, Salahuddin2007UltraST}. For example, high-pass filtering can eliminate slow movement/baseline changes, and hardware designs often include analog low-pass stages to suppress DC and low-frequency artifact. Nonetheless, simple filtering may not suffice when motion noise lies in-band \cite{Reddy2009UseOF}. Therefore researchers use more advanced techniques: adaptive filters that use accelerometer references, wavelet denoising, or machine learning to distinguish the pulse waveform from artifacts \cite{Raghuram2010, Biswas2019CorNETDL, Ibam2026MultimodalDL}. Compared with adaptive filtering approaches such as Least Mean Squares filtering or wavelet denoising, the present pipeline takes a simpler artifact-rejection approach rather than attempting to reconstruct corrupted PPG segments. LMS and wavelet-based methods can be effective when the artifact structure is sufficiently represented in a reference signal and when a usable physiological waveform remains present. In the AffectaPocket setting, however, tactile squeezing and inconsistent grip may not only add motion noise but can also temporarily degrade sensor-skin coupling or occlude blood flow. In such cases, reconstruction-based filtering may be less reliable because the underlying optical pulse signal is partially absent. We therefore prioritized a transparent IMU- and confidence-based pipeline that rejects high-motion samples and smooths post-motion recovery. This choice improves interpretability and is computationally simple for a future embedded robot implementation, but it also reduces temporal coverage and should be compared directly against LMS, wavelet, and learning-based denoising in future work.

\section{Method and Study Design}
The study utilized \textit{AffectaPocket}, as depicted in Figure \ref{fig:affecta}, a pocket-sized ubiquitous companion robot ($7\,\mathrm{cm} \times 4\,\mathrm{cm} \times 3\,\mathrm{cm}$) originally designed to assist children in managing anxiety through tactile interaction \cite{Frederiksen2024TowardAP}. The device features a 3D-printed anthropomorphic shell housing an ESP32 microcontroller, which manages capacitive touch sensors, articulated button inputs on the ``arms'' and ``feet,'' and a linear resonant actuator (LRA) for haptic feedback. While the robot's primary interaction involves a tactile rhythm-matching game designed to divert attention during anxious moments, the hardware architecture was upgraded for this experiment to enable physiological monitoring. We integrated a photoplethysmography (PPG) sensor board (MAXREFDES117 \cite{AnalogDevicesMAXREFDES117}) directly into the robot's chassis to capture blood volume pulse data during holding, alongside an internal Inertial Measurement Unit (IMU) to track device orientation and acceleration.  By comparing the robot's output against a Polar Verity Sense optical heart-rate sensor \cite{polar}, we assessed the impact of motion artifacts on data quality and the effectiveness of multi-stage filtering. The study evaluates heart-rate estimation rather than continuous emotional-state tracking or anxiety detection.

\subsection{Research Hypotheses}
The investigation was guided by five primary hypotheses:

\begin{itemize}
    \item \textbf{H1:} The unfiltered but smoothed data gathered from the robot will deviate from the reference sensor measurements by more than 40\%.
    \item \textbf{H2:} Removing data segments where accelerometer-based filtering (utilizing rolling standard deviation) suggests high movement will improve the accuracy of the original unfiltered data.
    \item \textbf{H3:} Introducing additional filtering based on signal confidence levels will bring the heart-rate measurements within a 25\% margin of the reference measurements.
    \item \textbf{H4:} The filtered data will be statistically equivalent to the reference sensor using Two One-Sided Tests (TOST) for equivalence within $\pm 5$ BPM bounds.
    \item \textbf{H5:} The filtered data will significantly improve precision by lowering the difference for paired samples to a statistically significant extent.
\end{itemize}

\begin{figure*}[h]
\centering
\includegraphics[width=1.00\textwidth]{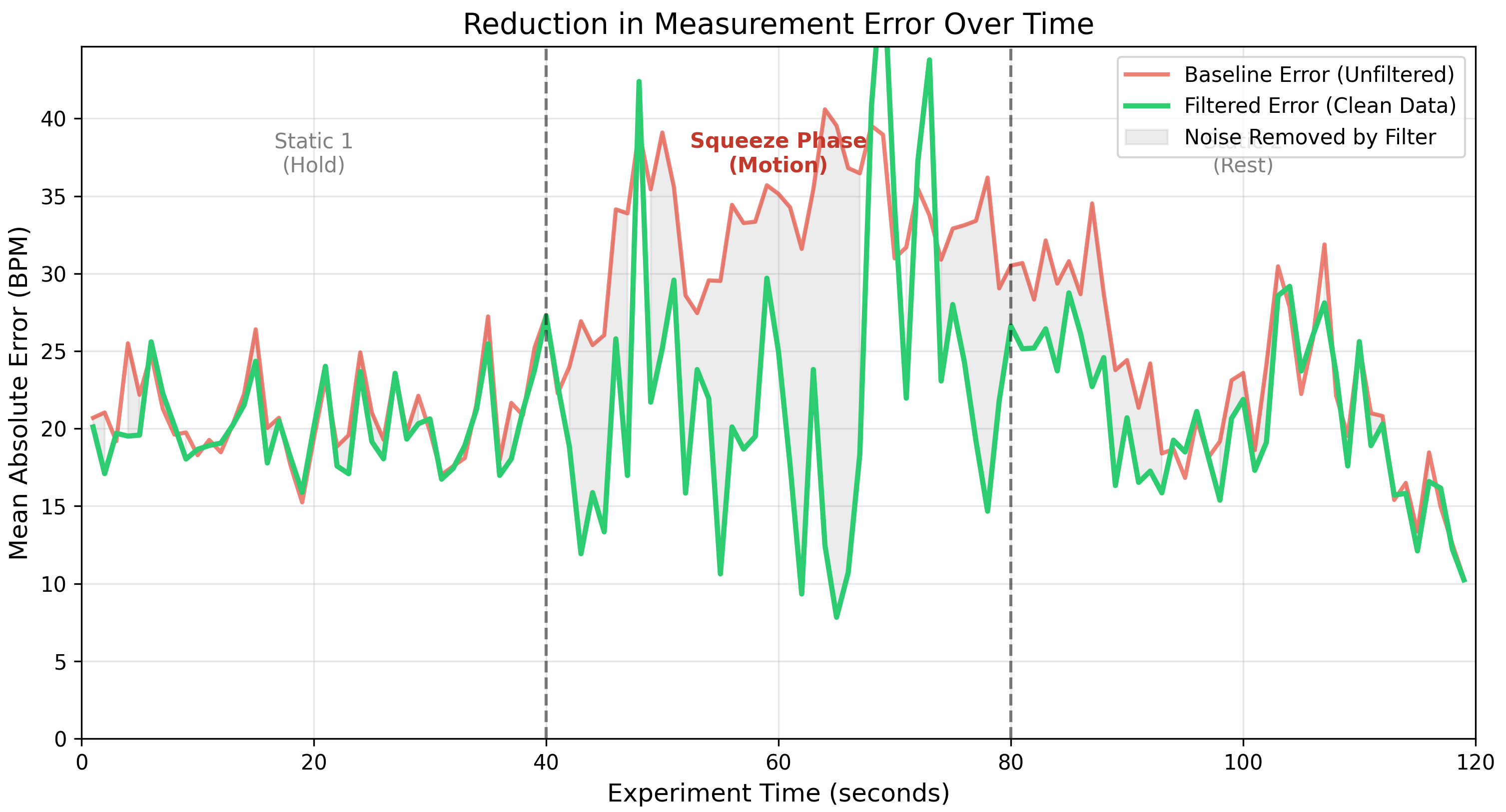}
\caption{Mean absolute error over time before and after applying IMU-based filtering and confidence-based smoothing. The red series depicts the baseline error from the unfiltered \textit{AffectaPocket} signal relative to the Polar Verity Sense reference sensor, while the green series depicts the error after filtering. Gray regions indicate samples removed by the motion filter}
\label{fig:filter_efficiency}
\end{figure*}

\subsection{Participants}
We included a total of 26 participants in this study. The demographic composition consisted of four female participants and 22 male participants. The age distribution was primarily centered in the 20–30 age group, which accounted for 23 participants. The remaining participants included two individuals in the 10–20 age group (Subject 2 and Subject 4) and one individual in the 30–40 age group (Subject 21). The participants were recruited at the university and they were compensated with a soft drink for their participation.

\subsection{Experimental Setup}
The study was conducted in a secluded test room at the university to ensure a controlled environment. Each participant session lasted for a total of 2 minutes.

Before the study began, participants were asked to provide their age group and gender. Following this, participants were instructed to hold the AffectaPocket robot in one hand, the side randomly assigned by the experimenter. The Polar Verity Sense optical heart-rate sensor was worn simultaneously according to the manufacturer's recommended placement. The robot and polar sensor recorded heart-rate estimates concurrently throughout the 120-second session. Because the reference device is itself an optical heart-rate sensor rather than electrocardiography, we treat it as a practical reference sensor for evaluating the embedded robot prototype. The study lasted 120 seconds, which was divided into three distinct 40-second phases to test the sensors under different motion conditions:

\begin{enumerate}
    \item \textbf{Initial Steady Phase (0–40sec):} Participants held the Affecta device steadily with minimal movement and were instructed not to press any buttons.
    \item \textbf{Interaction Phase (40–80sec):} Participants were instructed to press the side buttons of the device to induce movement. They were not specifically told to squeeze the device, resulting in varied interaction styles; some participants used their fingers to press buttons while others squeezed the chassis.
    \item \textbf{Final Steady Phase (80–120sec):} Participants returned to holding the device steadily with minimal movement and no intentional interaction.
\end{enumerate}

\subsection{Data Processing}
Data alignment was performed using the starting timestamps of the Polar Verity Sense sensor as the primary reference, while the AffectaPocket robot recorded its own internal timestamps for synchronization. Rows with heart-rate measurements outside the biologically plausible range of 40--200 BPM were dropped.

\subsubsection{Motion Rejection Algorithm}
The Affecta's onboard Inertial Measurement Unit (IMU) provided raw $x$, $y$, and $z$ acceleration data. To identify and reject motion artifacts, this 3-axis data was converted into a square-root magnitude:
\begin{equation}
    Mag = \sqrt{x^2 + y^2 + z^2}
\end{equation}
We then applied a rolling standard deviation to this magnitude to detect high-movement segments. 

\begin{equation}
    \text{Filter Value}_t = \sigma(\text{Magnitude}_t, \dots, \text{Magnitude}_{t-1s})
\end{equation}

The IMU and heart-rate data processing was not performed on-device. All IMU and heart-rate data were sent via Wi-Fi to a server and processed offline after collection.

\subsubsection{Signal Refinement}

To address the physiological signal recovery latency observed following mechanical artifacts, we implemented a \textit{Confidence Filter} that categorizes data quality based on the time elapsed since the last detected motion event. Unlike binary motion filtering, which re-enables raw data collection immediately upon the cessation of movement, this strategy enforces a stabilization period where the signal is treated with caution to account for sensor re-coupling. The filter logic is defined as follows:

\begin{equation}
    HR(t) = 
    \begin{cases} 
    \text{Discard (NaN)} & \text{if } \sigma_{\text{acc}}(t) > 400 \\
    \text{Median}(HR_{t-w}, HR_{t}) & \text{if } t - t_{\text{last\_move}} \le 5.0\text{s} \\
    HR_{raw}(t) & \text{if } t - t_{\text{last\_move}} > 5.0\text{s}
    \end{cases}
\end{equation}

\noindent where $\sigma_{\text{acc}}(t)$ denotes the rolling standard deviation of the accelerometer magnitude and $t_{\text{last\_move}}$ represents the timestamp of the most recent motion event. The acceleration-variance threshold of 400 and the 5-second recovery window were selected as pragmatic parameters based on inspection of pilot recordings and the temporal behavior of the PPG signal after visible movement events. The threshold was chosen to identify rapid device motion without removing all minor hand adjustments, while the recovery window was intended to cover the period in which the optical sensor often re-coupled with the skin after movement or pressure changes. These values should therefore be interpreted as prototype-level hyperparameters rather than optimized universal settings.

Because the first filtering stage discards high-motion samples rather than reconstructing them, the pipeline necessarily trades temporal coverage for improved accuracy. This is especially relevant during tactile interaction, where discarded samples may correspond to behaviorally meaningful events such as squeezing, button pressing, or stress-related manipulation of the robot. 

During the 5-second recovery window, the signal was processed using a rolling median filter ($w=5s$) to smooth transient recovery artifacts while avoiding immediate reactivation of unstable raw readings.
\section{Results}

The PPG sensor was evaluated against the reference signal using a baseline of $N=17,135$ biologically valid samples, defined as heart-rate measurements between 40 and 200 BPM. The initial hypothesis \textbf{H1} stated that unfiltered data would deviate by more than 40\% from the reference due to mechanical noise. However, analysis of the baseline dataset revealed a Mean Absolute Percentage Error (MAPE) of $29.14\% \pm 23.12\%$ and a mean bias of $-1.38 \pm 32.41$ BPM. Although the raw signal exhibited high variance, the average relative error was significantly lower than the anticipated threshold, leading to the rejection of Hypothesis \textbf{H1}.

\begin{table*}[ht]
\centering
\caption{Summary of robot PPG sensor performance showing the cumulative improvement of the filtering strategies. Performance is evaluated using MAPE (Mean Absolute Percentage Error), representing the average deviation from the Polar Verity Sense reference signal, and Corr. ($r$), the Pearson correlation coefficient indicating the strength of the linear relationship between the two datasets.}
\label{tab:results_summary}
\resizebox{\textwidth}{!}{%
\begin{tabular}{lcccc}
\toprule
\textbf{Processing Stage} & \textbf{Sample Size ($N$)} & \textbf{MAPE (\%)} & \textbf{Corr. ($r$)} & \textbf{Hypothesis} \\ \midrule
1. Baseline (Unfiltered) & 17,135 & $29.14 \pm 23.12$ & 0.165 & \textbf{H1}: Rejected \\
2. Motion Filtered Only & 9,974 & 24.20 & 0.245 & \textbf{H2}: Supported \\
3. Motion + Confidence Filtered & 9,974 & $22.72 \pm 20.61$ & 0.253 & \textbf{H3}: Supported \\ \bottomrule
\end{tabular}%
}
\end{table*}

\subsection{Impact of Motion Artifacts}
Hypothesis \textbf{H2} suggested that filtering segments with high accelerometer variance would improve data quality. Applying the Subtle Movement Filter (rolling standard deviation $> 400$) reduced the dataset to $N=9,974$ samples. This exclusion decreased the MAPE from $29.14\%$ to $24.20\%$ and reduced the Root Mean Square Error (RMSE) from $32.43$ to $27.80$ BPM. Furthermore, the correlation with the reference sensor improved from $r=0.165$ to $r=0.245$. These results confirm that motion variance is a significant source of error, supporting Hypothesis \textbf{H2}. Figure \ref{fig:filter_efficiency} visualizes this effect over the 120-second session, showing that the filtering pipeline removes high-motion segments during the interaction phase and reduces the error trajectory relative to the unfiltered baseline.
Adding the confidence-based filtering strategy, which smoothed the signal during the 5-second post-motion recovery window, further reduced MAPE of $22.72\% \pm 20.61\%$. This performance represents a robust accuracy level well within the targeted threshold, supporting Hypothesis \textbf{H3}, stating that introducing confidence-based filtering would align measurements within 25\% of the reference measurement. However, this improvement should be interpreted as improved accuracy over retained and smoothed samples, not as uninterrupted physiological coverage during all interaction moments.

\subsection{Statistical Equivalence and Significance of Improvement}
We hypothesized in \textbf{H4} that the filtered data would be statistically equivalent to the reference signal within $\pm 5$ BPM bounds. Two One-Sided Tests (TOST) performed on the final filtered set indicated statistical equivalence within the predefined $\pm 5$ BPM bounds ($p=0.003$, $p < 0.05$). The 90\% confidence interval for the mean bias was $[-4.72, -3.86]$ BPM, falling entirely within the equivalence bounds of $[-5, 5]$ BPM. This supports equivalence of the mean filtered heart-rate estimates within the predefined $\pm 5$ BPM bounds, but does not imply that individual second-by-second measurements are sufficiently precise for instantaneous closed-loop feedback.

Finally, Hypothesis \textbf{H5} proposed that the filtering strategy would significantly improve precision across paired samples. A paired samples $t$-test conducted on the mean MAPE per subject ($N=26$) compared the Baseline and Smart Recovery conditions. The test revealed a statistically significant improvement ($t(24)=-3.122$, $p=0.0023$), confirming that the reduction in error was consistent across participants and not attributable to random chance, thus supporting \textbf{H5}. T(24) was used as one participant's data was too sparse after filtering.

\subsection{Single measurement prediction analysis}
For analyzing how the system handles shorter duration observations and predictions on single measurements, the heart rate (HR) data was categorized into three physiological zones: Low ($<75$ BPM), Medium ($75-95$ BPM), and High ($>95$ BPM). The classification analysis across all filtered samples can be seen in Figure \ref{fig:confusion_matrix_corrected}, and yielded an overall accuracy of $48\%$. The system demonstrated the highest sensitivity for the Low category with a recall of $71.2\%$ ($2119/2975$ samples), suggesting that during resting or low-intensity phases, the sensor coupling on the robot remains most stable. In contrast, the Medium and High categories faced significant challenges, with recall rates of $40.0\%$ ($1838/4597$) and $35.3\%$ ($849/2402$), respectively. A significant portion of the error in the Medium range was due to underestimation, with $39.5\%$ of those samples being misclassified as Low. Similarly, High HR measurements were frequently misidentified as Low ($37.6\%$) or Medium ($27.1\%$).

\section{Discussion}

\subsection{Statistical Equivalence vs. Real-Time Precision}

A critical finding of this study is the discrepancy between the high statistical equivalence of the filtered data and its limited utility for real-time classification. While the Two One-Sided Tests (TOST) confirmed that the robot’s measurements were statistically equivalent to the reference sensor, with a  mean bias within 5 BPM, this metric primarily reflects the accuracy of the system over time rather than its instantaneous precision. The low bias indicates that the robot does not systematically over- or underestimate heart rate; however, the high standard deviation of the error reveals significant variance in second-by-second readings.

\begin{figure}[h!]
\centering
\includegraphics[width=0.50\textwidth]{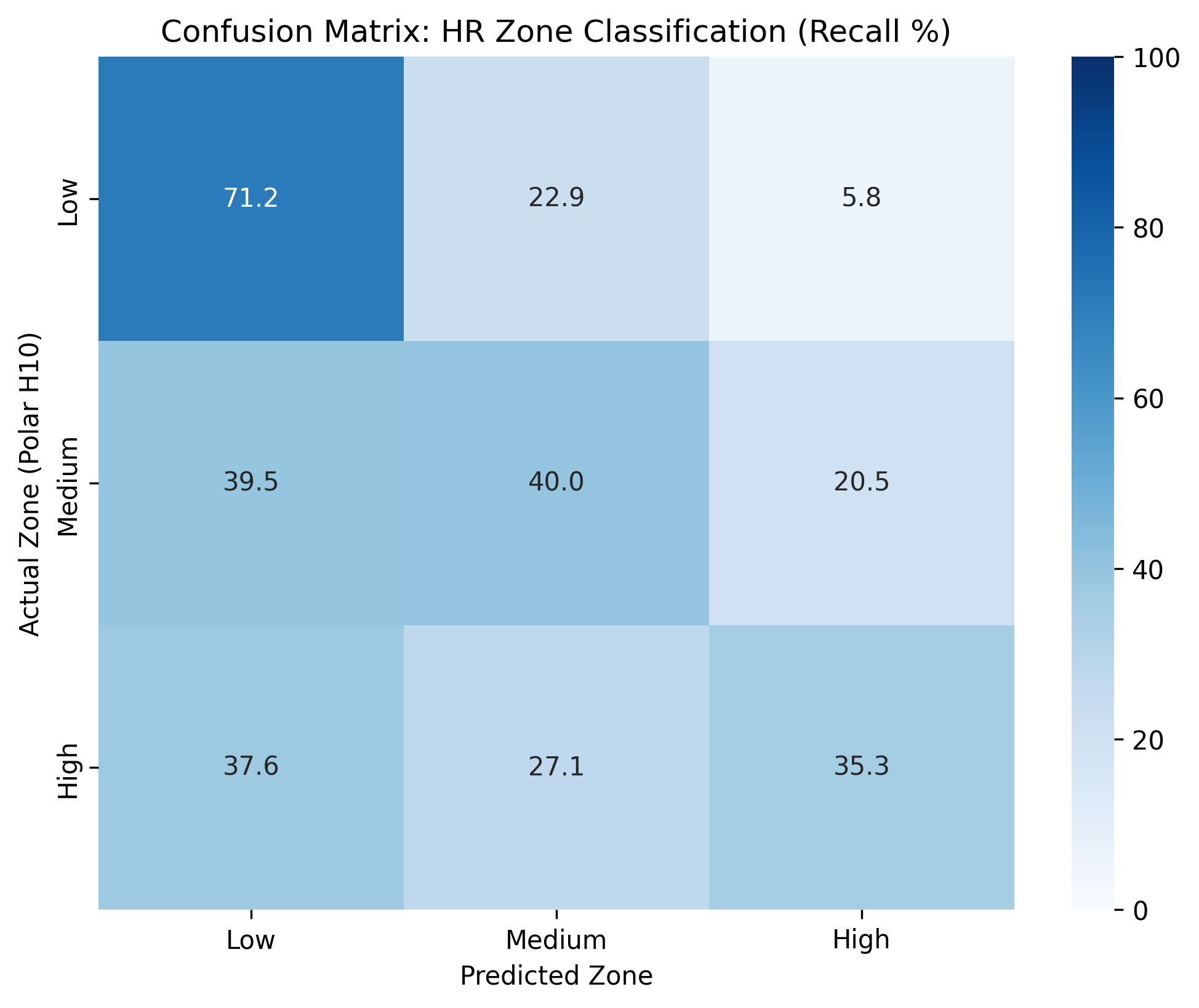}
\caption{Confusion Matrix for HR Zone Classification. The rows represent the actual heart rate zones determined by the Polar reference sensor (\textbf{Low}: $<75$ BPM, \textbf{Medium}: $75-95$ BPM, \textbf{High}: $>95$ BPM), while the columns represent the predicted zones. The diagonal values indicate the classification accuracy for each category: 71.2\% for Low, 40.0\% for Medium, and 35.3\% for High. The off-diagonal cells display the distribution of misclassifications.}
\label{fig:confusion_matrix_corrected}
\end{figure}

This results in a set of noisy data points that, while centered on the reference measurements, frequently deviates across the binary classification threshold (83 BPM). Consequently, although the system is reliable for monitoring average physiological trends over longer durations (e.g., 10-minute windows), its high variance compromises its ability to accurately classify instantaneous heart rate states (Low vs. High), yielding only a 48\% classification accuracy. The system’s underestimation of high heart rates, including misclassification as low in 37.6\% of instances (as shown in the confusion matrix in Figure \ref{fig:confusion_matrix_corrected}), likely stems from squeeze pressure momentarily occluding blood flow, a common issue with photoplethysmogram (PPG) sensors during tactile interaction.
This suggests that while the current filtering pipeline effectively removes extreme mechanical artifacts over sustained measurements (estimated to need $>$ 34sec), further signal smoothing or a windowed voting mechanism would be required to stabilize the signal for reliable short-duration real-time state detection within this context.
These results indicate that the current prototype should not be used as the sole trigger for real-time stress adaptation. In its present form, AffectaPocket can only support slow-changing, trend-level biofeedback, such as adapting interaction patterns after sustained changes in heart rate over a longer window.

\subsection{Generalizability to Children With Anxiety}

The motivating use case for \textit{AffectaPocket} is anxiety support for children, but the present evaluation was conducted primarily with university-age adults, and the sample was predominantly male. This limits the external validity of the results. Children may differ from adults in hand size, grip force, and the ways they manipulate a pocket-sized robot during stress. In addition, anxious or distressed use may involve more irregular squeezing and fidgeting than the controlled interaction phase used here.

Accordingly, the present study should be interpreted as an engineering feasibility evaluation rather than a validation of affective support for anxious children. Future work should evaluate the system with the intended child population.

\subsection{Coverage Loss Across Interaction Phases}

The motion-rejection stage reduced the dataset from 17,135 biologically valid samples to 9,974 retained samples, corresponding to an overall coverage loss of approximately 41.8\%. This confirms that the gain in accuracy comes with a  loss of temporal coverage. Because the experimental protocol separated the session into initial steady, interaction, and final steady phases, a phase-specific analysis would be important for determining whether the discarded samples are concentrated during the interaction phase.

\subsection{Hardware Enhancements for Ubiquitous Bio-Sensing}
While the current prototype utilizes a single optical PPG sensor, this configuration is highly susceptible to variations in placement and pressure, limitations that are magnified in ubiquitous computing contexts where users may interact with the device from unpredictable angles or grasps. In contrast, modern consumer wearables typically employ multi-sensor arrays. Adopting a similar distributed array configuration would allow ubiquitous robots to capture viable signals regardless of how the user initiates contact, significantly increasing the probability of obtaining immediate physiological data.

Furthermore, the system's ability to reject artifacts could be significantly enhanced by integrating capacitive touch sensors immediately flanking the optical window. Unlike accelerometer-based filtering, which infers contact quality indirectly from motion, capacitive sensors would provide a direct, binary confirmation of skin proximity. This would allow the ubiquitous system to gate the PPG signal, processing data only when tight skin contact is physically verified. This approach, ensuring that every sample entering the algorithm is physiologically valid, would require minimal computational overhead while potentially having the largest impact on the accuracy of short-duration measurements in unstructured environments.

\section{Conclusion}

This paper investigated the feasibility of embedded physiological sensing in ubiquitous pocket-sized robots to eliminate the need for external wearables. We expanded the functionality of a ubiquitous robot to include an embedded PPG sensor and evaluated a motion-rejection filtering strategy designed to mitigate the mechanical noise inherent in spontaneous handheld interactions. The system was tested in a within-subjects study with 26 participants, comparing the robot's embedded sensor performance against a Polar Verity Sense wrist worn reference sensor.

Our results demonstrate that the proposed IMU-based filtering significantly improved signal quality compared to the unfiltered baseline. The system achieved a final error rate that maintained statistical equivalence with the practical reference signal for extended interactions. However, the system's instability during short bursts indicates that single-sensor configurations lack the redundancy required for instantaneous readings. We conclude that while algorithmic filtering can enable reliable trend monitoring, achieving high-precision feedback during brief, spontaneous use will likely require hardware-level solutions, such as capacitive contact gating or multi-sensor arrays.

\balance

\bibliography{bibliography}
\bibliographystyle{IEEEtran}

\end{document}